\documentclass[journal]{IEEEtran}

\ifCLASSINFOpdf
\else
\fi
\usepackage{url}
\usepackage{verbatim}
\usepackage{amsmath,graphicx,amssymb}
\usepackage{physics}
\usepackage{multirow}
\usepackage{multicol}
\usepackage{accents}
\usepackage{subcaption}
\usepackage[ruled, vlined]{algorithm2e}
\usepackage{algorithmic}
\usepackage{xcolor}
\DeclareMathOperator*{\argmax}{arg\,max}

\newcommand\munderbar[1]{%
  \underaccent{\bar}{#1}}

\begin{document}
%
%
%
\title{
Normalizing Flow based Hidden Markov Models for \\ Classification of Speech Phones with Explainability
}
%

\author{Anubhab Ghosh, Antoine Honor\'e, Dong Liu, Gustav Eje Henter, Saikat Chatterjee \\
Digital Futures, and School of Electrical Engg. and Computer Sc., KTH Royal Institute of Technology, Sweden \\
anubhabg@kth.se, honore@kth.se, doli@kth.se, ghe@kth.se, sach@kth.se
}

\maketitle

\begin{abstract}

In pursuit of explainability, we develop generative models for sequential data. The proposed models provide state-of-the-art classification results and robust performance for speech phone classification. We combine modern neural networks (normalizing flows) and traditional generative models (hidden Markov models - HMMs). Normalizing flow-based mixture models (NMMs) are used to model the conditional probability distribution given the hidden state in the HMMs. Model parameters are learned through judicious combinations of time-tested Bayesian learning methods and contemporary neural network learning methods. We mainly combine expectation-maximization (EM) and mini-batch gradient descent. The proposed generative models can compute likelihood of a data and hence directly suitable for maximum-likelihood (ML) classification approach. 
Due to structural flexibility of HMMs, we can use different normalizing flow models. This leads to different types of HMMs providing diversity in data modeling capacity. The diversity provides an opportunity for easy decision fusion from different models. 
For a standard speech phone classification setup involving 39 phones (classes) and the TIMIT dataset, we show that the use of standard features called mel-frequency-cepstral-coeffcients (MFCCs), the proposed generative models, and the decision fusion together can achieve $86.6\%$ accuracy by generative training only. This result is close to  state-of-the-art results, for examples, $86.2\%$ accuracy of PyTorch-Kaldi toolkit \cite{Ravanelli2019}, and $85.1\%$ accuracy using light gated recurrent units \cite{Ravanelli2018}. We do not use any discriminative learning approach and related sophisticated features in this article.
\end{abstract}
%
\section{Introduction}
\label{sec:intro}

Neural networks based discriminative methods and generative models based maximum-likelihood (ML) methods are two main directions in pattern classification. ML classification is an optimal rule derived from Bayes minimum risk criterion under certain technical conditions. 

Neural networks and their deep versions in discriminative setups are successful to provide state-of-the-art classification performances across many applications.
Neural networks are data-driven, model-free, and are typically optimized for a pre-defined number of classes.
In many cases, structures of neural network based deep systems are not self-explanatory. It is difficult to understand individual roles of components or learning tricks in a deep system. This limitation in understanding is a major reason for lack of explainability.



On the other hand, model-based systems can be subjected to scrutiny and analysis for understanding. Generative models are typically model-based and can explain the process of data generation. If generative models can compute likelihood of a data point then they are suitable for ML classification. For example, a Gaussian mixture model (GMM) is a suitable generative model widely used for ML classification. Parameters of a GMM are learned from data using time-tested learning principles, such as expectation-maximization (EM). GMMs have been used in numerous applications with robust performance where data is corrupted. Our opinion is that  generative models with explainable data generation process, use of time-tested learning principles, and scope of robust performance for many potential applications provide a path towards explainable machine learning and trust.
In pursuit of explainability, we combine advantages of model-free neural networks and generative model-based ML classification systems. For modeling sequential data, we use hidden Markov models (HMMs) as a basic scheme. Then we develop new generative models by combining neural networks and HMMs. Performances of the proposed models are demonstrated for a speech recognition application in this article. While HMMs have been widely used for speech recognition  \cite{gales2008application}, they have been used in many other applications, such as handwriting recognition \cite{plamondon2000online}, activity recognition \cite{bao2004activity}, genomic signal processing \cite{anastassiou2001genomic}, transport forecasting \cite{piecyk2010forecasting}, etc. 

For speech recognition, the conditional probability distribution of a state in HMM is classically modelled using GMM. The GMM-HMM combination is a pure model-based system. GMM-HMMs are generative models that can be trained using time-tested probabilistic machine learning methods, such as EM, variational Bayes (VB) and Markov-chain-monte-carlo (MCMC). 
In spite of success, GMM-HMMs have limited modeling capability owing to the use of GMMs for representing complicated data manifolds.



We improve modeling capability of the conditional state distributions in an HMM. We use neural networks for the conditional state distributions.
There are recent advances in probability density modeling using neural networks called normalizing flows \cite{kobyzev2019normalizing}. Normalizing-flows are generative models where we can compute likelihood of a data point. Later, mixtures of normalizing flows were proposed and EM based training was shown in \cite{liu2020explicit}. A mixture of normalizing flows is a generative model that can handle multiple modes and manifolds of a data distribution. We refer to the proposed models as normalizing flow-based mixture models (NMMs). For our proposed models, we use NMMs as the conditional distributions of HMM states. That means, we replace time-tested GMMs by modern NMMs. The combination of NMM and HMM is referred to as NMM-HMM in contrast to GMM-HMM \cite{ghosh2020nmm}. 

There are several varieties of normalizing flows. We explore the use of RealNVP \cite{Dinh2019} and Glow  \cite{Kingma2018} models in our work, and we use succinct notations such as NVP-HMM and Glow-HMM to describe corresponding NMM variety in our work. There are other normalizing flow models, for example, auto-regressive (AR) flow \cite{Papamakarios2017}, that also fit in our development. We do not explore the other flow models due to high complexity.

This article builds on the prior works \cite{liu2020explicit,ghosh2020nmm,liu2019powering}. Our main technical contributions in this article are as follows. 
\begin{itemize}
    \item We develop NMM-HMM models (NVP-HMM and Glow-HMM) and show their better performance than GMM-HMM for a robust speech recognition problem under ML classification principle. We do not use any discriminative learning approach.
    \item We formulate  judicious combinations of EM and gradient search (back-propagation) that can learn the  parameters of NMM-HMMs.
    \item We show that appropriate combination of GMM-HMM and proposed NMM-HMMs using a simple decision fusion strategy yields state-of-the-art results. The decision fusion also provides robust classification performance. Our results conclusively shows generative model based ML classification is competitive. 
\end{itemize}
We performed extensive experiments for phone recognition using implementations based on Python, Kaldi and PyTorch. We used the TIMIT database and 39 phone classes as per standard practice in literature \cite{lopes2011phone,lee1989speaker}. 

\subsection{Relevant literature}

There is a growing literature on explainable machine learning, with a significant interest on (deep) neural networks.
Existing methods for explanations can be categorized in various ways, \cite{lipton2018mythos}, for example, \emph{intrinsic} and \emph{post-hoc} \cite{ribeiro2016should}. Intrinsic explainability is inherent in structures of methods, such as rule-based models, decision trees, linear models, and unfolding neural networks \cite{monga2019algorithm}. In contrast, a post-hoc method requires a second (simple) model to provide explanations about the original (complex) method. Therefore the proposed generative models in this article are close to intrinsic nature of explanability. There are other kind of approaches to provide
\emph{local} and \emph{global} explanations. A local explanation justifies a model or method's output for a specific input. A global explanation provides a justification on average performance of a model, independently of any particular input \cite{danilevsky2020survey}. There are techniques for explainability using visualization of neuronal activity in layers of a deep architecture, for examples, based on sparsity and heatmaps \cite{samek2019explainable, yosinski2015understanding, cadena2018diverse}. Another approach of explainability depends on explaining contributions of input features to output, referred to as feature importance \cite{lemhadri2019lassonet}. In the background of this literature survey, our proposed models combine model-based and data-driven methods for explaining data generation process. We also perform robustness study of our proposed models using corrupted data at varying noise conditions. Overall, our proposed models have connections with intrinsic explainability and global explainability.



We now provide a literature survey on speech phone classification.
For the TIMIT dataset, a list of phone recognition accuracy results on 39 phones is summarized in Github \cite{werweare}. 
According to the list, the best phone recognition accuracy is $86.2\%$ \cite{Ravanelli2019}. The work in \cite{Ravanelli2018} based on light gated recurrent units reported $85.1\%$ accuracy. The work \cite{toth2015phone} based on hierarchical convolutional deep maxout networks reported 83.5\% accuracy. Therefore, the ballpark accuracy for TIMIT phone recognition today is around $85-86\%$.

Prior to state-of-the-art results, an earlier attempt was to use restricted Boltzmann machines (RBMs) to form a deep belief network (DBN) that served as the acoustic models for HMMs \cite{mohamed2009deep}. 
The DBN used a softmax output layer and was discriminatively trained using backpropagation, achieving a phone recognition accuracy around $77\%$ on the TIMIT dataset.

Dynamical neural networks have also been used for speech recognition. Examples are recurrent neural networks (RNNs), long-short-term-memory networks (LSTMs) \cite{Graves2013}, and their gated recurrent unit based modifications. Attention mechanisms are also found to provide $82.4\%$ accuracy \cite{Chorowski2015}. Almost all these example works employ discriminative learning. 



There are end-to-end designs to achieve good performance for speech recognition \cite{Chorowski2015,lu2016segmental}. These methods typically learn appropriate features. In this article, we deliberately avoid feature learning or use of sophisticated features \cite{hermansky1992rasta, chatterjee2010auditory, koniaris2010selecting}. The reason is that we wish to test the power of proposed generative models solely. We perform all our experiments using time-tested features called MFCCs \cite{logan2000mel}. Development of MFCCs uses traditional speech signal processing knowledge.

\begin{figure*}[!ht]
    \centering
    \includegraphics[width=\textwidth]{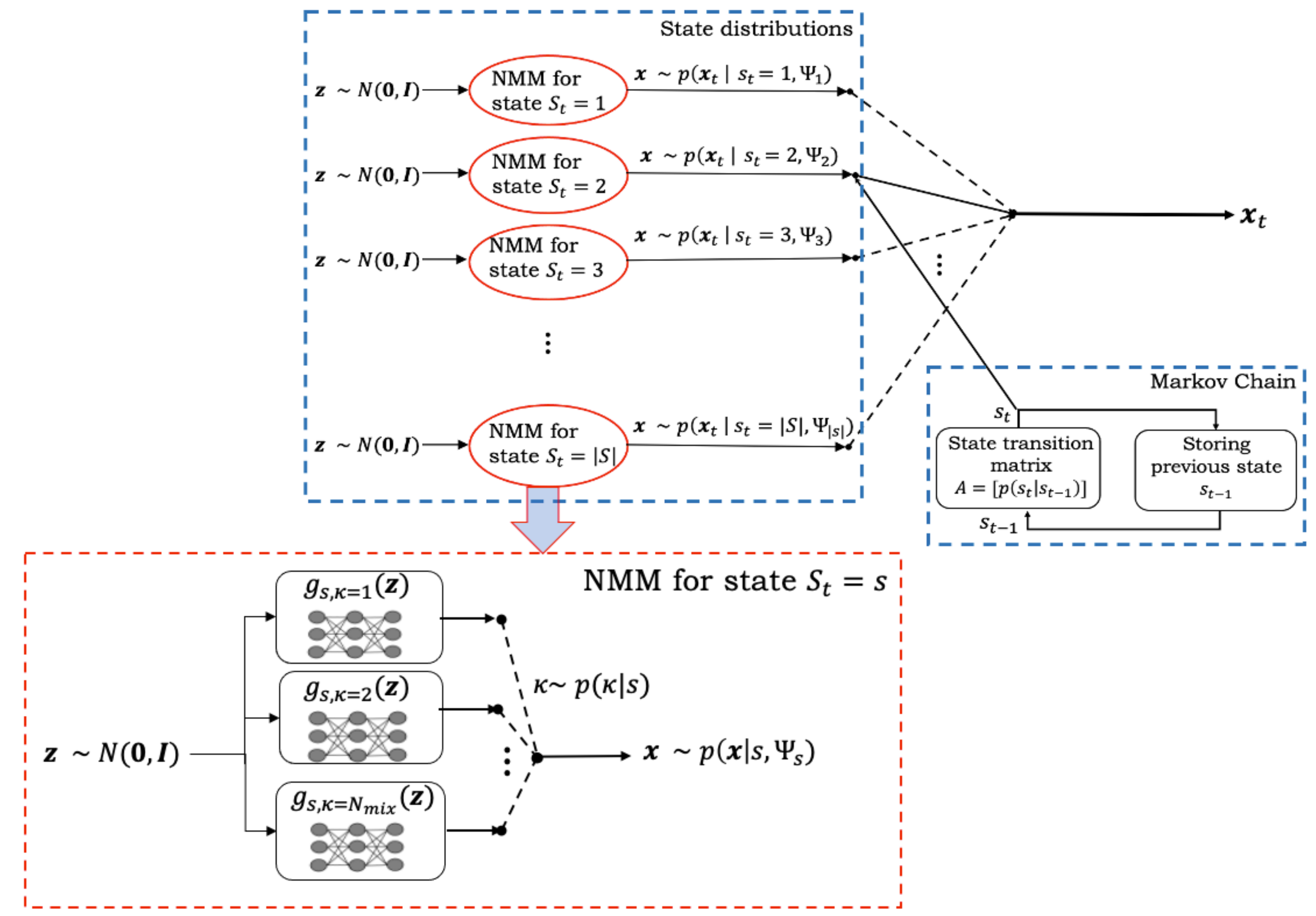}
    \caption{Schematic of the NMM-HMM architecture. The architecture inside a single NMM model for a given state $S_t = s$ is also shown in an expanded view at the bottom of the figure using red dotted line.}
    \label{fig:NMMHMMfig}
\end{figure*}
\section{NMM-HMM}
\label{sec:NMM-HMM}
Let the number of classes be $C$ for ML classification. We denote the generative model for the $c$'th class as $\mathbf{H}_c$. We denote a data sequence $\munderbar{\mathbf{x}} = \left[\mathbf{x}_1, \mathbf{x}_{2}, \ldots \mathbf{x}_{T}\right]^{\top}$ ($\top$ denotes the transpose operator), where $\mathbf{x}_{t} \in \mathbb{R}^{D}$ denotes the feature vector at time $t$, and $T$ denotes the sequence length. Then the time-tested ML-classification approach is
\begin{equation}
\label{MLeqn}
    \mathbf{H}_{c}^{\star} = \argmax_{H_{c}} p(\mathbf{\munderbar{x}} | \mathbf{H}_{c}),
\end{equation}
where $p(\mathbf{\munderbar{x}} | \mathbf{H}_c)$ denotes likelihood of the data sequence $\mathbf{\munderbar{x}}$ given the model $\mathbf{H}_c$. Here $\mathbf{H}_c$ can be a GMM-HMM or a NMM-HMM. We explain NMM-HMM below. 


\subsection{NMM-HMM architecture}
We now drop the subscript $c$ for notational clarity. 
An HMM is characterized by the following quantities: the set of hidden states of the HMM denoted by $\mathrm{\mathit{S}}$, the initial probability vector for the states of the HMM is denoted by $\mathbf{q}$, the state-transition probability matrix is denoted by $\mathbf{A}$, and the conditional probability distribution given the hidden state $s$ is denoted by $p\left(\mathbf{x}|s;\mathbf{\Psi}_{s}\right)$, where $\mathbf{\Psi}_{s}$ are the parameters. 
The conditional distribution given the hidden state $(s)$ in NMM-HMM is modelled as a weighted mixture of $N_{mix}$ density functions as
\begin{equation}
    p\left(\mathbf{x}|s;\mathbf{\Psi}_{s}\right) = \sum_{k=1}^{N_{mix}} \pi_{s,k} p\left(\mathbf{x}|s;\boldsymbol{\Phi}_{s,k}\right).
\label{eqn1}
\end{equation}
In \eqref{eqn1}, the weights $\lbrace \pi_{s,k} \rbrace$ denote the probability of drawing a given mixture component $k$ from a categorical distribution with $\pi_{s,k} = p\left(k | s;\mathbf{H}\right)$ and they satisfy $\sum_{k=1}^{N_{mix}} \pi_{s,k} = 1$. We use normalizing-flow based density function as the mixture component, denoted by $p\left(\mathbf{x}|s;\boldsymbol{\Phi}_{s,k}\right)$. A normalizing-flow function is a neural network $\mathbf{g}_{s,k} : \mathbb{R}^{D} \to \mathbb{R}^{D}$, such that
\begin{equation*}
\mathbf{x} = \mathbf{g}_{s,k}\left(\mathbf{z} \right),
\end{equation*}
where $\mathbf{z}$ is a $D$-dimensional latent variable that is drawn from a known prior distribution. We assume $p\left(\mathbf{z}\right)$ is a \textit{standard} isotropic Gaussian distribution  $\mathcal{N}(\mathbf{0},\mathbf{I})$. The function $\mathbf{g}_{s,k}$ is invertible. The inverse function $\mathbf{f}_{s,k}$ (or equivalently $\mathbf{g}^{-1}_{s,k}$) is also called \textit{normalizing} function. The use of the term `normalizing' is due to the reason that the signal $\mathbf{x}$ is generated from $\mathbf{z}$ sampled from a normal distribution $\mathcal{N}(\mathbf{0},\mathbf{I})$ \cite{kobyzev2019normalizing}. The inverse function is mathematically described as
\begin{equation*}
\mathbf{z} = \mathbf{f}_{s,k}\left(\mathbf{x} \right) \triangleq \mathbf{g}^{-1}_{s,k} (\mathbf{x}).
\end{equation*}
Using the change of variable formula in probability theory, we can calculate the likelihood for every mixture component as
\begin{equation}
\begin{aligned}
p\left(\mathbf{x}|s;\mathbf{\Phi}_{s,k}\right) &= p \left(\mathbf{f}_{s,k}\left(\mathbf{x} \right)\right) \abs{ \det \left( \mathbf{J}\left(\mathbf{f}_{s,k}\left(\mathbf{x}\right)\right) \right)}, \\
\end{aligned}
\label{eqn2}    
\end{equation}
where $\mathbf{J}\left(\mathbf{f}_{s,k}\left(\mathbf{x}\right)\right) = \left[\frac{\partial \mathbf{f}_{s,k}\left(\mathbf{x} \right) }{\partial \mathbf{x}}\right]$, denoting the Jacobian of the normalizing function $\mathbf{f}_{s,k}\left(\mathbf{x}\right)$. The mixture model in \eqref{eqn1} using normalizing flows, with each mixture component defined using \eqref{eqn2}, is referred to as Normalizing flow-based mixture model (NMM). The NMM model is
\begin{equation}
\begin{aligned}
\label{eqnNMM}
p\left(\mathbf{x}|s;\mathbf{\Psi}_{s}\right) &= \sum_{k=1}^{N_{mix}} \pi_{s,k} p \left(\mathbf{f}_{s,k}\left(\mathbf{x} \right)\right) \abs{ \det \left( \mathbf{J}\left(\mathbf{f}_{s,k}\left(\mathbf{x}\right)\right) \right)} \\
&= \sum_{k=1}^{N_{mix}} \pi_{s,k} p \left(\mathbf{g}_{s,k}^{-1}\left(\mathbf{x} \right)\right) \abs{ \det \left( \mathbf{J}\left(\mathbf{g}_{s,k}^{-1}\left(\mathbf{x}\right)\right) \right)}.
\end{aligned}
\end{equation}
The schematic of the NMM-HMM architecture is shown in Fig. \ref{fig:NMMHMMfig}. \\

It should be noted that NMMs satisfy certain specific properties in order to be incorporated into an existing probabilistic framework such as hidden Markov models. One important property is that the mapping defined by each mixture component from $\mathbf{z}$ to $\mathbf{x}$ is bijective. Thus, the mapping is one-to-one and an inverse mapping from $\mathbf{x}$ to $\mathbf{z}$ also exist. Also, the fact that the likelihood values can be computed exactly for an NMM enable similar mathematical treatment as a Gaussian mixture model (GMM). The log-determinant of the Jacobian $\mathbf{J}\left(\mathbf{f}_{s,k}\left(\mathbf{x}\right)\right)$ should be also easy to compute. Each mixture component of an NMM for a given state has its own set of parameters $\mathbf{\Phi}_{s,k}$, which require to be learned using back-propagation. This is illustrated by explanation of the learning problem and its solution in the subsequent section. 
\subsection{Learning of NMM-HMM parameters}
Let us assume we have $M$ number of training data sequences as $\{\munderbar{\mathbf{x}}^{(m)}\}_{m=1}^M$. Then the ML-based learning problem is:
\begin{equation}
\begin{aligned}
    \mathbf{H}^{\star} &= \argmax_{\mathbf{H} \in \mathcal{H}} \frac{1}{M} \prod_{m = 1}^{M} p \left(\mathbf{\munderbar{x}}^{(m)};\mathbf{H} \right)\\ 
    & \triangleq \argmax_{\mathbf{H} \in \mathcal{H}} \frac{1}{M} \log \left( \prod_{m = 1}^{M} p \left(\mathbf{\munderbar{x}}^{(m)};\mathbf{H} \right) \right)\\
    &= \argmax_{\mathbf{H} \in \mathcal{H}} \frac{1}{M} \sum_{m = 1}^{M} \log \left( p \left(\mathbf{\munderbar{x}}^{(m)};\mathbf{H} \right) \right),\\
\end{aligned}
\label{eqnmaxllh}
\end{equation}
where the possible hypothesis set of models is denoted by $\mathcal{H}$. 
The learning problem in \eqref{eqnmaxllh} can be solved using the well known expectation-maximization (EM) framework \cite{bishop2006pattern}. The expectation  step ("E-step") involves the calculation of the posterior probability distribution of hidden sequences of states $\mathbf{\munderbar{s}}$ and mixture components $\mathbf{\munderbar{k}}$, to obtain an expected value of the log-likelihood of the data sequence $\mathbf{\munderbar{x}}$ under the current model parameters $\mathbf{{H}}^{old}$. This is formulated as a cost function $\mathcal{L}\left(\mathbf{H} ; \mathbf{H}^{old}\right)$ and is shown in \eqref{eqnEstep} as
\begin{equation}
\mathcal{L}\left(\mathbf{H} ; \mathbf{H}^{old}\right) = \mathbf{E}_{p\left( \mathbf{\munderbar{s}}, \mathbf{\munderbar{k}} | \mathbf{\munderbar{x}}; \mathbf{H}^{old}\right)} \log \left( p\left( \mathbf{\munderbar{x}}, \mathbf{\munderbar{s}},  \mathbf{\munderbar{k}}; \mathbf{H}\right) \right).
\label{eqnEstep}    
\end{equation}
The second step, known as the maximisation step ("M-step"), consists of finding the model $\mathbf{H}^{\star}$ that maximizes the expected log-likelihood computed in \eqref{eqnEstep}. This can be decomposed into three separate maximization problems as
\begin{multline}
\max_{\mathbf{H}} \mathcal{L}\left(\mathbf{H} ; \mathbf{H}^{old} \right) \\
= \max_{\mathbf{q}} \mathcal{L}\left(\mathbf{q}; \mathbf{H}^{old} \right) + \max_{\mathbf{A}} \mathcal{L}\left(\mathbf{A}; \mathbf{H}^{old} \right) + 
\max_{\mathbf{\Psi}} \mathcal{L}\left(\mathbf{\Psi}; \mathbf{H}^{old} \right),
\label{eqnMStep}
\end{multline}
where,
\begin{equation}
\mathcal{L}\left(\mathbf{q}; \mathbf{H}^{old}\right) = \mathbf{E}_{p\left( \mathbf{\munderbar{s}} | \mathbf{\munderbar{x}}; \mathbf{H}^{old}\right)} \log \left( p\left( s_{1} ; \mathbf{H}\right) \right),
\label{eqnMStep1}    
\end{equation}
\begin{equation}
\mathcal{L}\left(\mathbf{A}; \mathbf{H}^{old}\right) = \mathbf{E}_{p\left( \mathbf{\munderbar{s}} | \mathbf{\munderbar{x}}; \mathbf{H}^{old}\right)} \sum_{t=2}^{T} \log \left( p\left( s_{t} | s_{t-1} ; \mathbf{H}\right) \right),
\label{eqnMStep2}    
\end{equation}
\begin{equation}
\mathcal{L}\left(\mathbf{\Psi}; \mathbf{H}^{old}\right) = \mathbf{E}_{p\left( \mathbf{\munderbar{s}}, \mathbf{\munderbar{k}} | \mathbf{\munderbar{x}}; \mathbf{H}^{old}\right)} \log \left( p\left( \mathbf{\munderbar{x}}, \mathbf{\munderbar{k}} | \mathbf{\munderbar{s}} ; \mathbf{H}\right) \right).
\label{eqnMStep3}    
\end{equation}
The maximisation problems described in \eqref{eqnMStep1}, \eqref{eqnMStep2} can be solved using standard EM forward-backward algorithm to compute the posterior distribution effectively, explained in \cite{bishop2006pattern}. For solving the maximisation problem in \eqref{eqnMStep3}, we need to maximise the log-likelihood with respect to the mixture of weights and the set of parameters of the flow models. This would require the computation of the log-determinant for the appropriate normalizing flow model. For solution of \eqref{eqnMStep3}, the problem can be broken down into two sub-problems, i.e. learning the set of mixture of weights $\mathbf{\Pi} = \lbrace \mathbf{\pi_{s}} | s \in S \rbrace $, and the learning the set of flow model parameters $\mathbf{\Phi} = \lbrace \mathbf{\phi_{s}} | s \in S \rbrace $. This is shown as
\begin{equation}
\max_{\mathbf{\Psi}} \mathcal{L}\left(\mathbf{\Psi} ; \mathbf{H}^{old} \right) \\
= \max_{\mathbf{\Pi}} \mathcal{L}\left(\mathbf{\Pi}; \mathbf{H}^{old} \right) + \max_{\mathbf{\Phi}} \mathcal{L}\left(\mathbf{\Phi}; \mathbf{H}^{old} \right).
\label{eqn15}    
\end{equation}
The problem of learning the mixture of weights can be solved using a simple Lagrangian formulation while problem of learning the flow model parameters can be solved by using the results of the change of variable formula derived in \eqref{eqn2} as
\begin{equation}
\begin{aligned}
\mathcal{L}\left(\mathbf{\Phi}; \mathbf{H}^{old} \right) &= \mathbf{E}_{p\left( \mathbf{\munderbar{s}}, \mathbf{\munderbar{k}} | \mathbf{\munderbar{x}}; \mathbf{H}^{old}\right)} \log \left( p\left( \mathbf{\munderbar{x}} | \mathbf{\munderbar{s}} , \mathbf{\munderbar{k}} ; \mathbf{H}\right) \right) \\
&= \mathbf{E}_{p\left(\mathbf{\munderbar{s}}, \mathbf{\munderbar{k}} | \mathbf{\munderbar{x}}; \mathbf{H}^{old} \right)} [ \log \left( p\left( \mathbf{f}_{s,k}\left(\mathbf{\munderbar{x}} \right) \right) \right)\\
&\quad + \log \left(\abs{ \det \left(\nabla \mathbf{f}_{s,k} \right)} \right) ],
\label{eqn16}
\end{aligned}
\end{equation}
The first term on the right hand side of \eqref{eqn16} is an expectation computed over the log-probability of latent data derived using the inverse function $\mathbf{f}_{s,k}$, and the second term is the result of the log-determinant of the Jacobian that is computed for the type of normalizing flow used. 
In the subsequent section, we explain the flow models that we used for modeling, and the computation of the log-determinant of the Jacobian $\left(\log \left(\abs{ \det \left(\nabla \mathbf{f}_{s,k} \right)} \right)\right)$.

\subsection{RealNVP flow}
We refer to a normalizing flow-based HMM (NMM-HMM) using RealNVP flow as NVP-HMM. The signal flow for a flow model having $L$ layers can be illustrated as
\begin{equation} \label{eqnnvp1}
    \mathbf{z}=\mathbf{h_0} \underset{\mathbf{f}_{s,k}^{[1]}}{\stackrel{\mathbf{g}_{s,k}^{[1]}}{\rightleftharpoons}} \mathbf{h_1} \underset{\mathbf{f}_{s,k}^{[2]}}{\stackrel{\mathbf{g}_{s,k}^{[2]}}{\rightleftharpoons}} \mathbf{h_2} 
    \underset{\mathbf{f}_{s,k}^{[3]}}{\stackrel{\mathbf{g}_{s,k}^{[3]}}{\rightleftharpoons}} \mathbf{h_3} \hdots 
    \underset{\mathbf{f}_{s,k}^{[L]}}{\stackrel{\mathbf{g}_{s,k}^{[L]}}{\rightleftharpoons}} \mathbf{h_L} = \mathbf{x},
\end{equation}
where $\mathbf{f}_{s,k}^{[l]}$ denotes the $l^{th}$ layer of the network $\mathbf{f}_{s,k}$, and each such $\mathbf{f}_{s,k}^{[l]}$ is invertible. One of the first variety of normalizing flows have been proposed in \cite{Dinh2015}, and different varieties discussed elaborately in \cite{kobyzev2019normalizing}. Different flow model architectures may have different kinds of \textit{coupling} between two successive layers in the network. An example can be obtained from the architecture called RealNVP flow that is discussed in \cite{Dinh2019}. To illustrate a small section of the mapping from the data space to the latent space, let us consider the input feature at the $l^{th}$ layer denoted by $\mathbf{h}_{l}$. This feature is mapped to $\mathbf{h}_{l-1}$ using the function $\mathbf{f}_{s,k}$. At every layer the $D$-dimensional input feature is split into two disjoint parts. Let us assume the features are $\left[\mathbf{h}_{l,1:d}, \mathbf{h}_{l,d+1:D}\right]^{T}$ (where $d$ denotes the number of components in the first sub part). The relation is
\begin{equation}
\begin{aligned}
\mathbf{h}_{l,1:d} &= \mathbf{h}_{l-1, 1:d}, \\
\mathbf{h}_{l,d+1:D} &= \mathbf{h}_{l-1, d+1:D} \odot \exp\left(\mathbf{s} \left(\mathbf{h}_{l-1,1:d} \right) \right) + \mathbf{t} \left(\mathbf{h}_{l-1,1:d}\right),\\
\end{aligned}
\label{eqnnvp2}
\end{equation}
where the symbol $\odot$ denotes element-wise multiplications, $\mathbf{s}: \mathbb{R}^{d} \to \mathbb{R}^{D-d}$, $\mathbf{t}: \mathbb{R}^{d} \to \mathbb{R}^{D-d}$, with $\mathbf{s, t}$ being shallow feed-forward neural networks that differ only in the activation function for the last layer, which is a hyperbolic tangent (\texttt{tanh}) activation function for modeling the logarithm of the standard deviation and an identity activation for the translation parameter \cite{Dinh2019}. The inverse mapping from the data space to the latent space is defined as\\
\begin{equation}
\begin{aligned}
\mathbf{h}_{l-1,1:d} &= \mathbf{h}_{l,1:d}, \\
\mathbf{h}_{l-1,d+1:D} &= \left( \mathbf{h}_{l, d+1:D} - \mathbf{t} \left(\mathbf{h}_{l,1:d} \right) \right) \odot \exp\left(-\mathbf{s} \left(\mathbf{h}_{l,1:d}\right) \right), \\
\end{aligned}
\label{eqnnvp3}
\end{equation}
A single layer of the mapping described in \eqref{eqnnvp2}, \eqref{eqnnvp3}, is referred to as a \textit{coupling layer} \cite{Dinh2019}. For the flow model, the inverse mapping shown in \eqref{eqnnvp3} is computed for every layer, and the determinant of the Jacobian matrix is computed as the product of the layer-wise determinants of the Jacobian matrices as
\begin{equation}
    \det\left(\nabla \mathbf{f}_{s,k} \right) = \prod_{l=1}^{L} \det\left(\nabla \mathbf{f}_{s,k}^{[l]} \right),
\label{eqnnvp4}
\end{equation}
where each $\det\left(\nabla \mathbf{f}_{s,k}^{[l]} \right)$ is computed as
\begin{equation}
\begin{aligned}
\det\left(\nabla \mathbf{f}_{s,k}^{[l]} \right) &= \det \left(\begin{bmatrix}
                                                \mathbf{I}_{1:d} & \mathbf{0} \\
                                                \frac{\partial \mathbf{h}_{l-1,d+1:D}}{\partial \mathbf{h}_{l-1,1:d}} & diag\left(-\mathbf{s}\left(\mathbf{h}_{l,1:d}\right)\right)
                                                \end{bmatrix}
                                                \right) \\
                                    &= \det \left(diag\left(-\mathbf{s}\left(\mathbf{h}_{l,1:d}\right)\right) \right). \\
\end{aligned}
\label{eqnnvp5}
\end{equation}
In \eqref{eqnnvp5}, $\mathbf{I}_{1:d}$ denotes an identity matrix and $diag(\dotsc)$ denotes a diagonal matrix with the elements of the argument vector in the main diagonal. It is important to ensure an alternate ordering between the two parts of the signal $\left[\mathbf{h}_{l,1:d}, \mathbf{h}_{l,d+1:D}\right]^{T}$ (described in the affine coupling layer) in \eqref{eqnnvp2} so that the required transformation is achieved. If the mapping in the $l^{th}$ layer is defined in \eqref{eqnnvp2}, then the mapping in the $(l+1)^{th}$ layer is defined as
\begin{equation}
\begin{aligned}
\mathbf{h}_{l+1,1:d} &=  \mathbf{h}_{l, 1:d} \odot \exp\left(\mathbf{s} \left(\mathbf{h}_{l,d+1:D} \right) \right) + \mathbf{t} \left(\mathbf{h}_{l,d+1:D}\right),\\
\mathbf{h}_{l+1,d+1:D} &= \mathbf{h}_{l, d+1:D}.\\
\end{aligned}
\label{eqnnvp6}
\end{equation}
\subsection{Generative 1 $\times$ 1 convolution flow (Glow)}
A variation of the RealNVP model used for generating image signals is known as Glow (Generative 1 $\times$ 1 convolution based flow) \cite{Kingma2018}. Similar to RealNVP, a normalizing flow-based HMM (NMM-HMM) using Glow is referred to as Glow-HMM. It introduces two important modifications: an \textit{activation normalization} layer to standardise the activations at every flow step and an \textit{invertible convolution} layer as a generalization of the switching operation between channels. In its simplest form, a Glow model consists of layers that are referred to as \textit{flow steps} (analogous to $L$ layers in case of RealNVP). The signal flow for a Glow model having $K$ flow-steps can be illustrated as
\begin{equation} \label{eqnglow1}
    \mathbf{z}=\mathbf{h_0} \underset{\mathbf{f^{'}}_{s,k}^{[1]}}{\stackrel{\mathbf{g^{'}}_{s,k}^{[1]}}{\rightleftharpoons}} \mathbf{h_1} \underset{\mathbf{f^{'}}_{s,k}^{[2]}}{\stackrel{\mathbf{g^{'}}_{s,k}^{[2]}}{\rightleftharpoons}} \mathbf{h_2} 
    \underset{\mathbf{f^{'}}_{s,k}^{[3]}}{\stackrel{\mathbf{g^{'}}_{s,k}^{[3]}}{\rightleftharpoons}} \mathbf{h_3} \hdots 
    \underset{\mathbf{f^{'}}_{s,k}^{[K]}}{\stackrel{\mathbf{g^{'}}_{s,k}^{[K]}}{\rightleftharpoons}} \mathbf{h_K} = \mathbf{x}.
\end{equation}
Each $i^{th}$ flow-step is composed of the three layers as 
\begin{equation}
\label{eqnglowcompose}
    \mathbf{f^{'}}_{s,k}^{[i]} = \mathbf{f^{'}_1}_{s,k}^{[i]} \circ \mathbf{f^{'}_2}_{s,k}^{[i]} \circ \mathbf{f^{'}_3}_{s,k}^{[i]}.
\end{equation}
In \eqref{eqnglowcompose},  $\mathbf{f^{'}_1}_{s,k}^{[i]}$ denotes the activation normalization layer, 
$\mathbf{f^{'}_2}_{s,k}^{[i]}$ denotes the invertible convolution layer, and $\mathbf{f^{'}_3}_{s,k}^{[i]}$ denotes the affine coupling layer. We refer to the \textit{forward mapping} as the transformation from the data space to the latent space, and the \textit{reverse mapping} as the transformation from the latent space to the data space. The log-determinant is usually computed for the forward mapping at each of the constituent layers shown in \eqref{eqnglowcompose}. Each of the individual layers of a single flow-step are described as follows:

\paragraph{Activation normalization}
It is assumed that an input vector is represented as $\mathbf{x}$ and is of the shape $(N_{b} \times N_{c} \times N_{s})$, where $N_b$ denotes the batch-size, $N_c$ denotes the number of channels and $N_s$ denotes the samples. In the following equation, $\mathbf{x}$ can be the starting input vector or the activation at any intermediate hidden layer in the flow network. 
The forward, reverse functions and log-determinant are
\begin{equation}
\label{eqn_actnorm_method2}
\begin{aligned}
    \mathbf{z} &= \left( \mathbf{x} - \boldsymbol{\mu}_{bias} \right) \odot \exp \left( \boldsymbol{\sigma}_{scale}^{\star} \right)   \hspace{0.3in} \left(\text{Forward direction}\right), \\
    \mathbf{x} &= \mathbf{z} \odot \exp\left(-\boldsymbol{\sigma}_{scale}^{\star} \right) + \boldsymbol{\mu}_{bias} \hspace{0.3in} \left(\text{Reverse direction}\right),\\
    \mathbf{J}_{1} &= \sum_{\text{channel dim.}} \boldsymbol{\sigma}_{scale}^{\star} \hspace{0.3in} \left(\text{Log-determinant}\right).
\end{aligned}
\end{equation}
In \eqref{eqn_actnorm_method2}, the values of $\boldsymbol{\mu}_{bias}$ and $\boldsymbol{\sigma}_{scale}$ are calculated as the mean and the logarithm of the standard deviation along each of the channel dimensions. 
\paragraph{Invertible 1 x 1 convolution}
The invertible 1 x 1 convolution is said to be a generalization of the permutation along the channel dimensions \cite{Kingma2018}. Any convolution operation can be thought of as a matrix multiplication. 
The forward, reverse and log-determinant calculation are shown in \eqref{invconv_method1} as
\begin{equation}
 \label{invconv_method1}
 \begin{aligned}
 \mathbf{z} &= \mathbf{W} \mathbf{x} \hspace{0.3 in} \text{(Forward direction)},\\
 \mathbf{x} &= \mathbf{W}^{-1} \mathbf{z} \hspace{0.3 in} \text{(Reverse direction)},\\
 \mathbf{J}_{2} &= \log\left(\abs{\det\left(\mathbf{W}\right)}\right) \hspace{0.3 in} \text{(Log-determinant)}.
 \end{aligned}
 \end{equation}
\paragraph{Affine coupling layer}
The affine coupling layer used in Glow is very similar to that in RealNVP. It is consists of a neural network function $\mathbf{N()}$ that is usually a \textit{shallow} feed-forward network using convolutional layers. 
The computations and log-determinant calculation in the forward direction are
\begin{equation}
\label{affine_glow1_method}
\begin{aligned}
\mathbf{x}_{a}, \mathbf{x}_{b} &= \mathbf{\texttt{split}\left(x\right)},\\
\log\left(\boldsymbol{\sigma}\right), \boldsymbol{\mu} &= \mathbf{N}(\mathbf{x}_{b}),\\
\boldsymbol{\sigma} &= \exp\left(\log \left(\boldsymbol{\sigma}\right)\right), \\
\mathbf{z}_{a} &= \left( \mathbf{x}_{a} + \boldsymbol{\mu} \right) \odot \boldsymbol{\sigma},\\
\mathbf{z}_{b} &= \mathbf{x}_{b}, \\
\mathbf{z} &= \texttt{concat}\left(\mathbf{z}_{a}, \mathbf{z}_{b} \right), \\
\mathbf{J}_{3} &= \sum_{\text{channel dim.}} \boldsymbol{\sigma}_{scale} \hspace{0.3in} \left(\text{Log-determinant}\right).
\end{aligned}
\end{equation}
The design as per \eqref{affine_glow1_method} indicates that $\log\left(\boldsymbol{\sigma}\right)$ is modelled as an logarithm of the inverse of the scaling parameter and $\boldsymbol{\mu}$ is modelled as negative of the translation parameter. 
The only difference in integrating the Glow model with the HMM comes in the formulation of the log-determinant similar to \eqref{eqnnvp4} for RealNVP flow. Using the individual determinant calculations from \eqref{eqn_actnorm_method2}, \eqref{invconv_method1}, \eqref{affine_glow1_method},  we can calculate $\log \left(\abs{ \det \left(\nabla \mathbf{f}_{s,k} \right)} \right)$ as
\begin{equation}
\label{eqnglowJ1}
\begin{aligned}
\log \left(\abs{ \det \left(\nabla \mathbf{f^{'}}_{s,k} \right)} \right) &= \sum_{l=1}^{K} \log \left(\abs{ \det \left(\nabla \mathbf{f^{'}}_{s,k}^{[l]}\right)}\right) \\
&= \sum_{l=1}^{K} \left(\mathbf{J}_{1}^{[l]}  + \mathbf{J}_{2}^{[l]} + \mathbf{J}_{3}^{[l]} \right). \\
\end{aligned}
\end{equation}
In \eqref{eqnglowJ1}, $K$ refers to the number of \textit{flow-steps} (without considering a multi-scale flow architecture i.e. assuming $L=1$). So once the log-determinant is calculated, we can substitute the value and calculate the cost function as in \eqref{eqn16}. 
\subsection{Training algorithm}
\label{sec:training_algorithm}
The essential steps of the learning procedure are described in psuedocode in Algorithm \ref{algo1}.
The algorithm involves training a flow-based model using Expectation-Maximization (EM) algorithm \cite{bishop2006pattern} and Back-propagation. It involves two essential steps: \\

\begin{itemize}

    \item The first step which involves calculation of the posterior distribution $p\left( \mathbf{\munderbar{s}}, \mathbf{\munderbar{k}} | \mathbf{\munderbar{x}}; \mathbf{H}^{old}\right)$, and the loss function $\mathcal{L}\left(\mathbf{\Phi}; \mathbf{H}^{old} \right)$ involving the parameters of the output distribution. However, unlike common generative models like GMM-HMM, the NMM-HMM models require to be trained before they can be used, without which the sample statistics would no longer be stationary. So, there is an inner loop to train the flow-models using mini-batch gradient descent with Adam \cite{kingma2014adam}. Training continues until a convergence criterion is satisfied or the maximum number of iterations is reached. The convergence criterion $\mathcal{C^{*}}$ was defined as
    \begin{equation*}
        \mathcal{C^{*}} : \left[ \left(\frac{\abs{\Lambda_{current} - \Lambda_{prev}}}{\abs{\Lambda_{prev}}} < \Delta \right) == \mathbf{True}\right]
    \end{equation*}
    If $\mathcal{C^{*}}$ was satisfied for a chosen number of times ($n^{\star}$) in succession, the training was stopped, else the model continued to be trained until the maximum number of training iterations ($N_{max}$) was reached. The convergence threshold $(\Delta)$ was decided carefully through experimentation. \\
    
    \item The second step which consists of updating the Markov-chain parameters $\mathbf{q}, \mathbf{A}$ and the mixture model weights $\mathbf{\Pi}$ were updated using standard optimization techniques of HMM training. 
    
\end{itemize}

These two steps are iteratively continued until the likelihood maximization reaches a local optimum.

\begin{algorithm}
\SetAlgoLined
 \KwIn{Dataset $\{\munderbar{\mathbf{x}}^{(m)}\}_{m=1}^M$, initial model parameters}
\KwResult{Optimized model parameters: $\mathbf{q}, \mathbf{A}, \mathbf{\Pi}, \mathbf{\Phi}$}
\textbf{Set training parameters}: learning rate $(\eta)$, no. of samples in a mini-batch ($R_b$), max. number of training epochs $(N_{max})$, convergence threshold ($\Delta$), no. of convergence steps ($n^{*}$), previous negative log-likelihood ($\Lambda_{prev}$), and  current negative log-likelihood ($\Lambda_{current}$)\\
\textbf{Initialization} of models: $\mathbf{H^{old}}$, $\mathbf{H}$:
where, $\mathbf{H} = \lbrace \mathbf{q}, \mathbf{A}, S, p\left(\mathbf{x}|s;\mathbf{\Psi}_{s}\right) \rbrace$,
and set $\mathbf{H^{old}} \gets \mathbf{H}$\\

 \While{$\mathbf{H}$ has not converged}{
  \While{ $\left(n \leq N_{max}\right)$ or $\mathcal{C^{*}}$ is satisfied}{
  Input a mini-batch from the dataset as $\lbrace \mathbf{\munderbar{x}}^{r} \rbrace ^{R_{b}}_{r = 1}$, with batch-size $R_{b}$ \\
  Compute the posterior $p\left( \mathbf{\munderbar{s}}^{r}, \mathbf{\munderbar{k}}^{r} | \mathbf{\munderbar{x}}^{r}; \mathbf{H}^{old}\right)$ 
  and the cost function $\mathcal{L}\left(\mathbf{\Phi}; \mathbf{H}^{old} \right)$ as per \eqref{eqnnvp5} or \eqref{eqnglowJ1} \text{(depending on the type of flow model)}  \\
  Compute $\partial \mathbf{\Phi}$ by optimising $\mathcal{L}\left(\mathbf{\Phi}; \mathbf{H}^{old} \right)$ (Mini-batch gradient descent using Adam \cite{kingma2014adam})\\
  Update $\mathbf{\Phi} \gets \mathbf{\Phi} + \eta \cdot \partial \mathbf{\Phi}$\\
  Update $n \gets n + 1$\\
  }
  Update $\mathbf{q} \gets \argmax_{\mathbf{q}}\mathcal{L}\left(\mathbf{q}; \mathbf{H}^{old} \right)$ \\
  Update $\mathbf{A} \gets \argmax_{\mathbf{A}}\mathcal{L}\left(\mathbf{A}; \mathbf{H}^{old} \right)$ \\
  Update $\mathbf{\Pi} \gets \argmax_{\mathbf{\Pi}}\mathcal{L}\left(\mathbf{\Pi}; \mathbf{H}^{old} \right)$ \\
  Update $\mathbf{H^{old}} \gets \mathbf{H}$\\
 }
 \caption{Learning algorithm for training NMM-HMMs}
 \label{algo1}
\end{algorithm}

\section{Experiments and Results}
\label{sec:experiments and results}
This section describes evaluation of NMM-HMM models (NVP-HMM and Glow-HMM) for phone recognition in comparison with GMM-HMM models. 

\subsection{Dataset, software and hardware}
The experiments were carried out using the TIMIT dataset \cite{lopes2011phone}. A phone is a distinct speech sound and is universal irrespective of the language under consideration. Phonemes characterize different phones in the context of a specific language. The TIMIT dataset is an English corpus, consists of spoken utterances labelled at the phoneme level. We use phones and phonemes interchangeably in this article. 
We use a `folded' set of 39 phonemes as per standard convention \cite{lee1989speaker}. 
The TIMIT dataset is comprised of two parts - a training set and a testing set, consisting of 4620 and 1680 utterances respectively. 
For creating noisy data, we used four varieties of additive noises from the NOISEX-92 dataset \cite{Varga1993}. 
The code for the NMM-HMM models (NVP-HMM and Glow-HMM) was written using PyTorch and run using GPU support. The code for the GMM-HMM model was written in Python using a package called \textit{hmmlearn} \cite{hmmlearn}. The codes are available at \url{https://github.com/anubhabghosh/genhmm}.

\subsection{Feature extraction}
From the speech utterances we compute MFCCs.
We used 13-dimensional MFCCs computed frame-by-frame with 25ms window length and 10 ms frame shift. As per standard practice we also compute dynamic features delta ($\Delta$) and double-delta ($\Delta\Delta$) coefficients from the MFCCs. So, in total, we use 39-dimensional feature vector concatenating MFCCs, $\Delta$ and $\Delta\Delta$. We performed standard mean and variance normalization.

\begin{table*}[!ht]
    \centering
    \caption{Test accuracy (in \%) for GMM-HMM and NMM-HMM models at varying number of mixture components $(N_{mix})$ in clean training and clean testing scenario}
    \resizebox{0.6\textwidth}{!}{\begin{tabular}{|*{7}{c|} }
            \hline
            \multirow{2}{*}{Model-Type}
            & \multicolumn{6}{c|}{No. of mixture components ($N_{mix}$)} \\
            \cline{2-7}
            & $N_{mix}$=1 & $N_{mix}$=3 & $N_{mix}$=5 & $N_{mix}$=10 & $N_{mix}$=15 & $N_{mix}$=20 \\
             \cline{1-7}
            GMM-HMM & 62.3 & 66.7 & 68.5 & 70.8 & 71.9 & 72.8 \\
            \hline
            NVP-HMM & 76.7 & 77.6 & - & - & - & - \\
            \hline
            Glow-HMM & 76.3 & - & - & - & - & - \\
            \hline
            \hline
        \end{tabular}}
    \label{tab:GMM_NVP_Glow_diff_components}
\end{table*}

\subsection{Training of models} 
For the GMM-HMM and NMM-HMM models, the number of hidden states $(s)$ was chosen between 3 and 5, depending on the mean length of speech signal (in terms of number of samples). The initial probability vector $\mathbf{q}$ was initialised as a vector of zeros, except the first state having probability one. The state transition probability matrix $\mathbf{A}$ was initialised as a upper triangular matrix. In case of GMM-HMMs, we use diagonal covariance matrices for the Gaussian components.   
For both NMM-HMM models, the conditional distribution for each state was formulated as in \eqref{eqn1} and \eqref{eqn2}. We experimentally find the number of mixture components $(N_{mix})$ for both the models. We also explain some additional, model-specific terms for NMM-HMMs. For NVP-HMM, we refer to a pair of consecutive coupling layers described in \eqref{eqnnvp1} and \eqref{eqnnvp2} as a \textit{flow block}. It was necessary to ensure that the signal in the $l^{th}$ layer would be alternated in the ${(l+1)}^{th}$ layer, so that there is a mixing of the signals between consecutive coupling layers. For Glow-HMM, we define a similar concept in terms of the parameters $K$ and $L$, which refers to the number of \textit{flow-steps} and number of \textit{layers of multi-scale flow} respectively. One \textit{flow-step} in the Glow model consists of three layers: activation normalization, invertible convolution, and an affine coupling layer (similar to RealNVP flow) shown in \eqref{eqnglowcompose}. We avoided the use of multi-scale flows and chose the value of $K$ through experiments. \\

We use the Adam optimizer for realizing mini-batch gradient descent in NMM-HMM models \cite{kingma2014adam}. The learning rate of gradient descent was experimentally set to be $\eta_{NVP} =4 \times 10^{-3}$ for NVP-HMM and $\eta_{Glow} = 1 \times 10^{-4}$ for Glow-HMM. Additionally, the learning rate was adaptively decreased by a step-decay to aid training for more number of iterations. The EM algorithm involved maximization of the log-likelihood (or minimization of the negative log-likelihood), and monitoring relative change of the log-likelihood as a convergence criterion (explained in Section \ref{sec:training_algorithm}).


\subsection{Performance results}
\label{sec:results}
We now show performance results of competing models for different training and testing conditions.

\subsubsection{Clean training and testing}
\label{sec:results1}
The performances of the NVP-HMM and Glow-HMM models were compared with the baseline GMM-HMM model for training and testing on clean data. The results are shown in Table \ref{tab:GMM_NVP_Glow_diff_components}. 
The GMM-HMM model was trained and tested on the clean data, for varying number of mixture components, i.e. $N_{mix} = \lbrace1, 3, 5, 10, 15, 20 \rbrace$, and the model with the best performance was chosen.  It was found that test accuracy for GMM-HMM increased with more number of mixture components. Based on the results, we decide to use $N_{mix}=20$ for all our GMM-HMM models. \\

    

We next consider NVP-HMM and it is found that $N_{mix}=3$ provides an improvement than $N_{mix}=1$. 
Further improvement by increasing number of components was challenging for us due to high computational complexity. Therefore we decide to use NVP-HMM using $N_{mix}=3$ components. 
Comparing results in Table  \ref{tab:GMM_NVP_Glow_diff_components}, we found that the NVP-HMM model with $N_{mix}=3$ mixture components outperforms the baseline GMM-HMM model with $N_{mix}=20$ components. The performance improvement is $4.8\%$. Here we mention that the NVP-HMM based ML-classification provides a similar classification accuracy for TIMIT phone recognition in comparison with discriminative training based DBN that had $77\%$ accuracy \cite{mohamed2009deep}. 

Similar experiments were also carried out for the Glow-HMM model. It was first necessary to decide upon the number of flow-steps $(K)$ required for the Glow model. As Glow-HMM is computationally complex, we used a smaller set of training data consisting of 5 classes (or phonemes) to decide the value of $K$. The result is shown in Table \ref{tab:Glow_diff_components}.
\begin{table}[!htbp]
    \centering
    \caption{Training and Test accuracy (in \%) for Glow-HMM ($N_{mix}=1, L=1$) at varying number of flow-steps (K) using the subset of clean data where we use ${5}$ classes}
    \begin{tabular}{|*{5}{c|} }
            \hline
            \multirow{2}{*}{Accuracy (in \%)}
            & \multicolumn{4}{c|}{No. of flow-steps ($K$)} \\
            \cline{2-5}
            & $K$=4 & $K$=8 & $K$=12 & $K$=16 \\
             \cline{1-5}
            Training & 90.6 & 94.9 & 96.9 & 96.1 \\
            \hline
            Testing & 89.7 & 93.9 & 96.0 & 95.89 \\
            \hline
            \hline
        \end{tabular}
    \label{tab:Glow_diff_components}
\end{table}
and we decide to use $K=12$ flow-steps and $L=1$ (no multi-scale flow). Also, it is worth noting that the accuracy values are high because it is relatively easier for the model when we use five classes instead of 39 classes. In order to achieve good performance, the use of \textit{weight normalization} on the weights of the convolutional layer in the affine coupling layer of the Glow model in each flow-step was helpful, and also normalizing the activations by their euclidean norm. The weight normalization approach was inspired by its use in the \textit{WaveGlow} model \cite{prenger2019waveglow}. We believe that use of normalization schemes ensures better convergence towards desired negative log-likelihood values during training of Glow-HMM. 
Glow-HMM provided 76.3 \% accuracy for 39 phones. This result is included in Table \ref{tab:GMM_NVP_Glow_diff_components} for a comparison. Simulations for Glow-HMM were only carried out using $N_{mix} = 1$. For a higher number of mixture components, it was computationally expensive for us to perform further experiments. \\

\begin{table}[!htbp]
    \centering
    \caption{Overall weighted Precision, Recall, and F1 Score (in \%) for three models in the clean training and clean testing scenario}
    \begin{tabular}{|c|c|c|c|}
    \hline
    \multirow{2}{*}{Model-Type} & \multicolumn{3}{c|}{Weighted Metrics} \\
    \cline{2-4}
    & Precision & Recall & F1-Score \\
    \cline{1-4}
    GMM-HMM &  75.4 & 72.5 & 73.2 \\
    \hline
    NVP-HMM &  77.5 & 77.0 & 76.6 \\
    \hline
    Glow-HMM &  78.6 & 76.4 & 74.8 \\
    \hline
    \hline
    \end{tabular}
    \label{tab:allmodelmetrics}
\end{table}

While we have provided accuracy values in Table \ref{tab:GMM_NVP_Glow_diff_components}, we also show other performance measures such as precision, recall and F1-score in Table \ref{tab:allmodelmetrics}. We computed class-wise precision, recall and F1-scores and combined them into a weighted metric based on the number of samples in each class. Precision is computed as the ratio of the number of correct positive predictions in that class to the total number of positive predictions for the class. Recall (or sensitivity) measures the fraction of correct positive predictions to the number of actual positive cases, while F1-score is a weighted mean of precision and recall for the given class. The results show that the Glow-HMM has the largest precision, which reflects its ability to distinguish true and false positives. Also, the recall value of Glow-HMM is similar to NVP-HMM, which indicates that the NMM-HMM models are better at differentiating false negatives than GMM-HMM. But, the F1-score reflects that NVP-HMM is equally capable like Glow-HMM for clean data.\\

\begin{table*}[!htbp]
    \centering
    \caption{Test accuracy (in \%) for clean and noisy testing conditions, where we use clean training. The performance drop is shown in parenthesis with respect to the clean train and clean test scenario as in Table \ref{tab:GMM_NVP_Glow_diff_components}}
    \setlength{\tabcolsep}{10pt}
    \resizebox{0.7\textwidth}{!}{\begin{tabular}{|*{5}{c|}}
            \hline
            \multicolumn{5}{|c|}{Clean data performance as a reference: GMM-HMM: 72.8, NVP-HMM: 77.6 and Glow-HMM: 76.3} \\
            \hline \hline
            \multirow{2}{*}{Type of Model}
            & \multicolumn{4}{c|}{SNR levels for different kinds of noises} \\
            \cline{2-5}
            & $SNR=25$dB & $SNR=20$dB & $SNR=15$dB & $SNR=10$dB \\
            \cline{2-5}
            & \multicolumn{4}{c|}{\texttt{white} noise} \\
            \hline
            \hline
            GMM-HMM & 55.6 (17.2) & 46.8 (26.0) & 36.8 (36.0) & 27.9 (44.9) \\
            \hline
            NVP-HMM & 67.1 (10.5) & 60.0 (17.6) & 49.4 (28.2) & 37.7 (39.9) \\
            \hline
            Glow-HMM & 67.9 (8.4) & 61.4 (14.9) & 53.6 (22.7) & 45.8 (30.5) \\
            \hline
            & \multicolumn{4}{c|}{\texttt{pink} noise} \\
            \hline
            \hline
            GMM-HMM & 59.9 (12.9) & 51.9 (20.9) & 42.3 (30.5) & 32.2 (40.6) \\
            \hline
            NVP-HMM & 69.3 (8.3) & 61.7 (15.9) & 48.6 (29.0) & 33.7 (43.9) \\
            \hline
            Glow-HMM & 69.4 (6.9) & 62.9 (13.4) & 54.9 (21.4) & 45.9 (30.4) \\
            \hline
            & \multicolumn{4}{c|}{\texttt{babble} noise} \\
            \hline
            \hline
            GMM-HMM & 65.7 (7.1) & 59.3 (13.5) & 49.3 (23.5) & 37.4 (35.4) \\
            \hline
            NVP-HMM & 70.7 (6.9) & 65.8 (11.8) & 56.2 (21.4) & 42.3 (35.3) \\
            \hline
            Glow-HMM & 72.4 (3.9) & 68.9 (7.4) & 63.1 (13.2) & 53.8 (22.5) \\
            \hline
            & \multicolumn{4}{c|}{\texttt{hfchannel} noise} \\
            \hline
            \hline
            GMM-HMM & 62.3 (10.5) & 54.4 (18.4) & 44.1 (28.7) & 33.3 (39.5) \\
            \hline
            NVP-HMM & 67.9 (9.7) & 63.4 (14.2) &  55.8 (21.8) & 44.9 (32.7) \\
            \hline
            Glow-HMM & 71.8 (4.5) & 67.9 (8.4) & 61.1 (15.2) & 51.9 (24.4) \\
            \hline
            \hline
            \end{tabular}}
    \label{tab:cl_tr_no_te_table_all}
\end{table*}

\subsubsection{Clean training and noisy testing - robustness test}
\label{sec:results2}
We now report results of experiments for checking the robustness of the NMM-HMM models relative to GMM-HMM in a mismatched training-testing scenario. We trained all the three models using clean (noise-less) data and tested them on noisy data. We used four types of noises: white, pink, babble and high frequency channel (labelled as `hfchannel') at different signal-to-noise-ratio (SNR) levels. We expect that a model robust to noise, would show a good performance as well as a graceful degradation in performance as the noise power increases (or equivalently SNR decreases). 
Table \ref{tab:cl_tr_no_te_table_all} shows the results. We observe that the performances of NMM-HMM models are better than GMM-HMM. We also note that Glow-HMM shows a better robustness to noise than NVP-HMM as exhibited by the graceful degradation of performance drop. \\

\vspace{-0.1cm}
\subsubsection{Noisy training and noisy testing - robustness test}
\begin{figure}[!htbp]
    \centering
    \includegraphics[width=3.5 in]{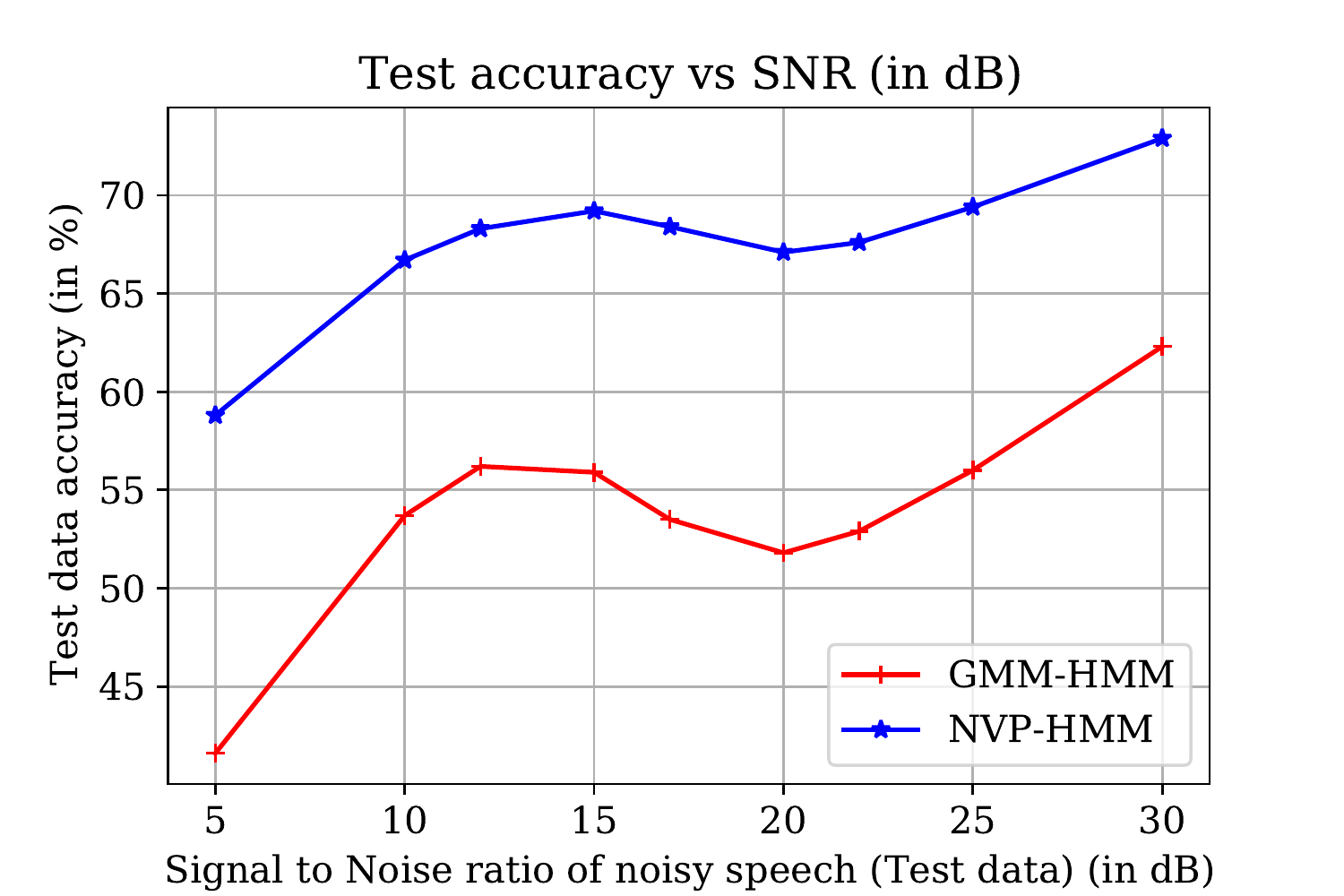}
    \caption{Robust performance for noisy training. Both models were trained on a combination of clean data and noisy data where we used white noise at 10dB SNR. Then the test is performed using noisy test data where white noise is used at varying SNR values.}
    \label{fig:MCT_plot}
\end{figure}

In this set of experiments, the GMM-HMM and NVP-HMM were trained using a dataset that is comprised of clean as well as noisy data, and then tested on varying noisy conditions. In the training dataset, the portion of the noisy data is generated by corrupting the clean data using white noise at 10dB SNR. Test results for varying SNR conditions (only white noise) are shown in Table~\ref{tab:no_tr_no_te_table_white}. 
The test performance was evaluated on the clean test set, as well as white noise corrupted test set at different SNR levels. 
It was hypothesized that there would be an increment in test accuracy for noisy conditions when compared to the respective results in Table \ref{tab:cl_tr_no_te_table_all}. The reason is the models have now been trained on noisy data. The results of Table \ref{tab:no_tr_no_te_table_white} are plotted in Fig. \ref{fig:MCT_plot} and we observe an interesting trend. Performance of both the models increases with SNR at first, then drops and then further increases. There is a peak closer to 10 dB SNR. The reason lies how the training and testing conditions match. Note that the training data has a portion that is corrupted by white noise at 10 dB SNR. From this experiment, it is clear that NVP-HMM is more robust than GMM-HMM. 

\begin{table*}[!htbp]
    \centering
    \caption{Test accuracy (in \%) for GMM-HMM and NVP-HMM using noisy training}
    \resizebox{0.8\textwidth}{!}{\begin{tabular}{|c|c|c|c|c|c|c|c|c|c|c|}
            \hline
            \multirow{2}{*}{Model-Type} &
            \multirow{2}{*}{Clean} &
            \multicolumn{9}{c|}{Diff. SNR levels for adding \texttt{white} noise}\\
            \cline{3-11}
            &  & 30dB & 25dB & 22dB & 20dB & 17dB & 15dB & 12dB & 10dB & 5dB\\
            \cline{1-11}
            GMM-HMM & 72.0 & 62.3 & 56.0 & 52.9 & 51.8 & 53.5 & 55.9 & 56.2 & 53.7 & 41.6\\
            \hline
            NVP-HMM & 76.8 & 72.9 & 69.4 & 67.6 & 67.1 & 68.4 & 69.2 & 68.3 & 66.7 & 58.8\\
            \hline
            \hline
        \end{tabular}}
    \label{tab:no_tr_no_te_table_white}
\end{table*}

\subsection{Improvements using decision fusion}
While we achieved promising performance using generative models based ML classification in the previous experiments, we are still away from the state-of-the-art performance $86.2\%$ accuracy \cite{Ravanelli2019}. Our main objective in this subsection is describing the engineering approach that can provide a performance close to state-of-the-art. Our hypothesis was that the different models behaved differently across phones (classes), and that developing an approach to combine their predictions would yield a better classifier. In this section, we describe our approach towards a \textit{decision fusion (or voting)} scheme using GMM-HMM, NVP-HMM and Glow-HMM. We use traditional speech processing knowledge for grouping phonemes as per different categories such as Vowels, Semi-vowels, Plosives, etc. We hypothesized that this would help us to investigate whether some trends in performance can be observed based on the given category of phonemes and  
the statistical aspect of data availability for the particular class. Before explaining the scheme, we explain two aspects.
\begin{itemize}
    \item \textit{Sample ratio:} For a phone class, sample ratio is computed as  
    \begin{equation*} \label{eqn_Ctrain}
        \mathrm{Sample \,\, ratio} = \frac{\text{\# tr. samples of the class}} {\text{\# tr. samples in the largest class }}
    \end{equation*}
    This ratio helps us to understand the role of data availability for classification performance. Here `largest class' means the `silence' class for which the number of training samples is the highest among all classes.
    \item \emph{Voting:} We use a majority voting principle. Among three models, if at-least two models are in agreement then we choose the agreed classification decision. Otherwise, all the three methods are in disagreement and then we choose a classification decision as a random choice among them. Note that voting is amenable to accommodate a growing number of classes.
\end{itemize}


\begin{figure}[!t]
    \centering
    \subcaptionbox{Vowels\label{figphones:vowel}}{\includegraphics[width=0.5\textwidth]{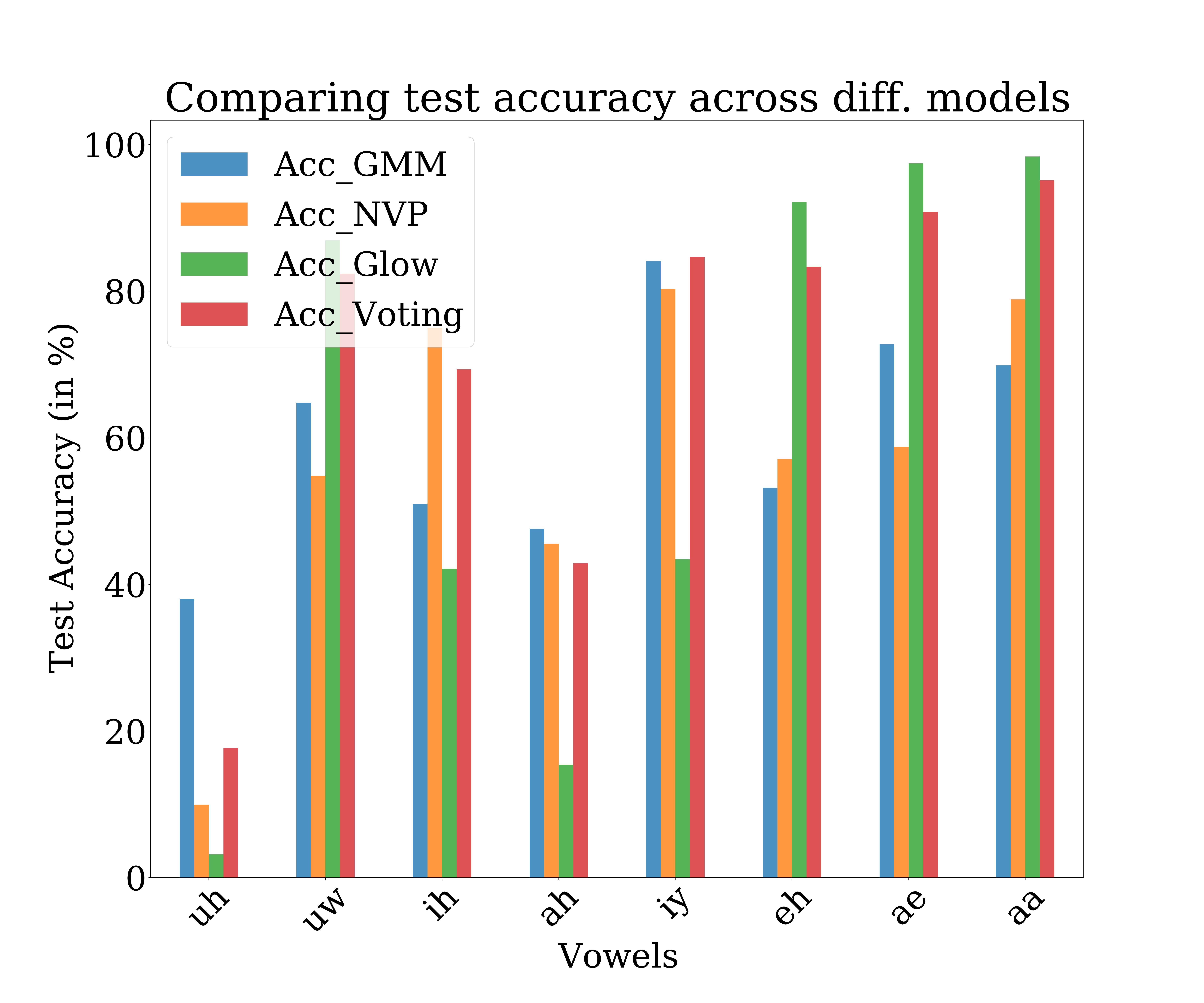}}
    \subcaptionbox{Fricatives\label{figphones:fricatives}}{\includegraphics[width=0.5\textwidth]{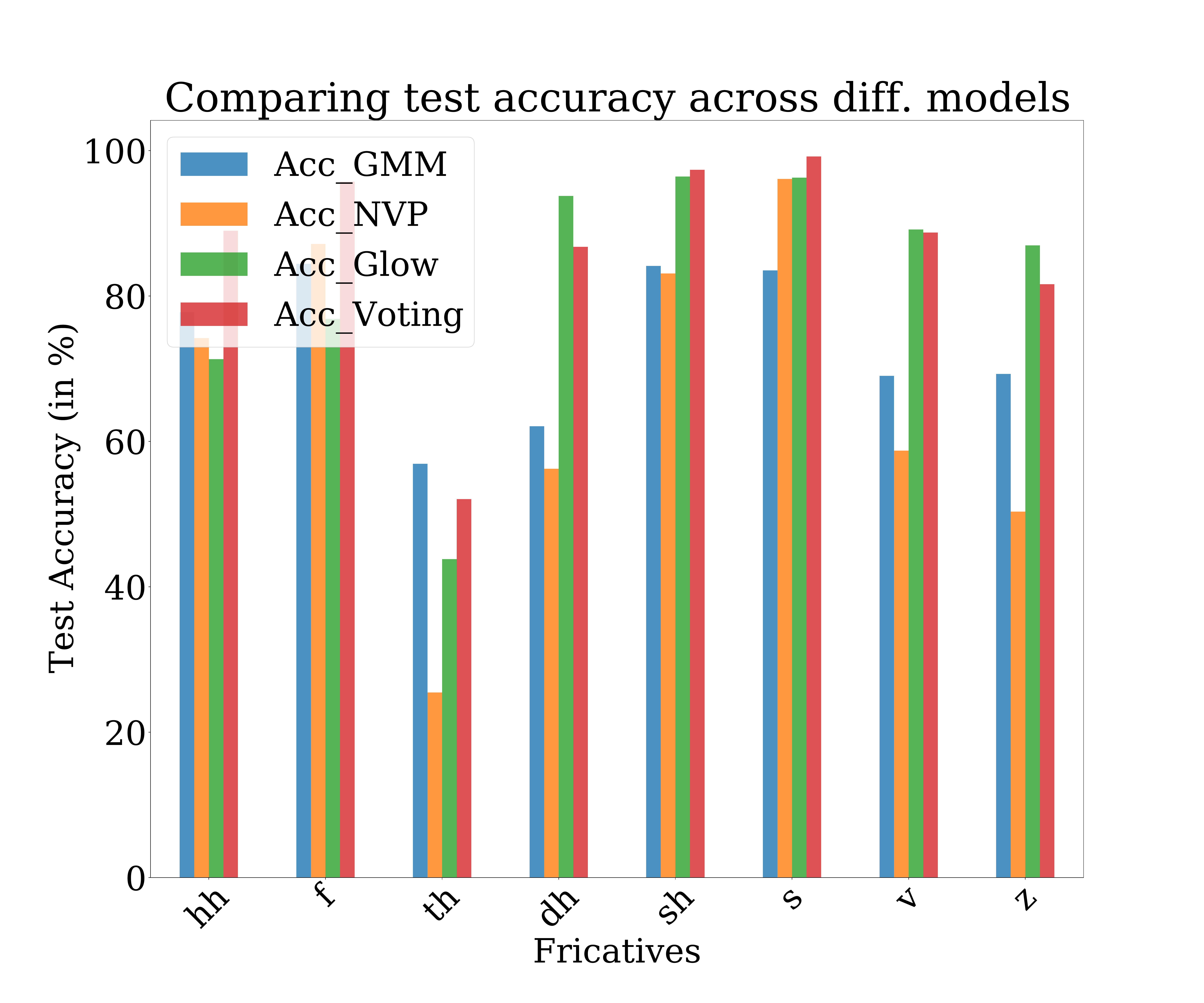}}
    \caption{Plot of test  accuracy for different models - GMM-HMM (Blue), NVP-HMM (Yellow), Glow-HMM (Green) -  against voting-based decision fusion (in Red) for vowels and fricatives}
    \label{fig:commachine_diff_phones}
\end{figure}
Using this voting scheme, we were able to achieve an accuracy $86.6\%$ on the test set, which is close to the state-of-the-art performance of $86.2\%$ in \cite{Ravanelli2019}. We now show class-wise accuracy results for each model in Table \ref{tab:voting_results}. From the table, we observe a significant variation in performances of the three models. We show few examples. For `aa' vowel, while GMM-HMM provides $69.9\%$ accuracy, the Glow-HMM provides $98.38\%$ accuracy. Then, for several  dipthongs, such as `ow', the Glow-HMM shows poor performance. For ease of visualization, we show variation in performance for vowels and fricatives in Fig.~\ref{fig:commachine_diff_phones}. This significant variation in performances is related to statistical diversity and helps the voting scheme to achieve $86.6\%$ accuracy. The $86.6\%$ accuracy is a significant improvement than the three models. \\

Our final experiment investigates the robustness of the voting based decision fusion approach. In this case, we have models that are trained using clean condition data and tested in noisy conditions. The results are shown in Table \ref{tab:cl_tr_no_te_table_all_voting}. In the table we compare the voting against the best performing model among the three models, which can be seen from the Table~\ref{tab:cl_tr_no_te_table_all}. We observe that the voting approach gracefully degrades with increasing noise power.

\begin{table*}[!htbp]
    \centering
    \caption{Test accuracy (in \%) for phones using voting based decision fusion against GMM-HMM, NVP-HMM and Glow-HMM. We used clean training and clean testing scenario.}
    \resizebox{\textwidth}{!}{\begin{tabular}{|*{8}{c|}}
            \hline
            Index &
            Phoneme &
            Type &
            Sample ratio (\%)
            & Acc. GMM - HMM
            & Acc. NVP - HMM
            & Acc. Glow - HMM
            & Acc. Voting \\
            \cline{1-8}
            \hline
            1  &      uh &    V &     1.49 &    38.01 &     9.95 &      3.17 &       17.65 \\
            \hline
            2  &      uw &    V &     6.87 &    64.80 &    54.80 &     86.93 &       82.40 \\
            \hline
            3  &      eh &    V &    10.75 &    53.19 &    57.08 &     92.15 &       83.33 \\
            \hline
            4  &      ae &    V &    11.15 &    72.78 &    58.78 &     97.44 &       90.83 \\
            \hline
            5  &      aa &    V &    16.75 &    69.90 &    78.90 &     98.38 &       95.11 \\
            \hline
            6  &      ah &    V &    17.55 &    47.59 &    45.54 &     15.41 &       42.89 \\
            \hline
            7  &      iy &    V &    19.39 &    84.13 &    80.30 &     43.43 &       84.69 \\
            \hline
            8  &      ih &    V &    38.20 &    50.95 &    74.99 &     42.14 &       69.32 \\
            \hline
            \hline
            9  &       y &   SV &     4.78 &    76.18 &    64.67 &     49.68 &       77.44 \\
            \hline
            10 &       w &   SV &     8.75 &    81.84 &    83.05 &     68.28 &       90.48 \\
            \hline
            11 &      er &   SV &    15.21 &    76.49 &    80.11 &     84.33 &       92.62 \\
            \hline
            12 &       r &   SV &    18.23 &    68.09 &    65.83 &     99.29 &       90.61 \\
            \hline
            13 &       l &   SV &    18.83 &    65.21 &    84.99 &     99.04 &       96.33 \\
            \hline
            \hline
            14 &      ch &    P &     2.29 &    72.97 &    54.44 &      1.54 &       51.35 \\
            \hline
            15 &      jh &    P &     3.37 &    69.09 &    63.17 &     75.54 &       82.80 \\
            \hline
            16 &       g &    P &     5.57 &    71.18 &    68.23 &     33.65 &       70.24 \\
            \hline
            17 &       b &    P &     5.59 &    74.28 &    66.04 &     64.92 &       83.15 \\
            \hline
            18 &       p &    P &     7.15 &    67.90 &    77.22 &     76.48 &       88.35 \\
            \hline
            19 &       d &    P &     9.51 &    68.21 &    71.61 &     91.23 &       90.73 \\
            \hline
            20 &       t &    P &    12.11 &    62.82 &    72.66 &     86.95 &       87.41 \\
            \hline
            21 &       k &    P &    13.54 &    77.28 &    83.86 &     62.88 &       88.89 \\
            \hline
            \hline
            22 &      ng &    N &     3.82 &    78.76 &    29.59 &     47.73 &       58.23 \\
            \hline
            23 &       m &    N &    11.22 &    77.17 &    78.38 &     62.44 &       84.76 \\
            \hline
            24 &       n &    N &    24.42 &    63.34 &    83.31 &     73.38 &       87.94 \\
            \hline
            \hline
            25 &      th &    F &     2.10 &    56.93 &    25.47 &     43.82 &       52.06 \\
            \hline
            26 &       v &    F &     5.56 &    69.01 &    58.73 &     89.15 &       88.73 \\
            \hline
            27 &      hh &    F &     5.89 &    77.79 &    74.21 &     71.31 &       88.97 \\
            \hline
            28 &       f &    F &     6.18 &    84.43 &    87.17 &     76.86 &       95.72 \\
            \hline
            29 &      sh &    F &     6.66 &    84.14 &    83.10 &     96.44 &       97.36 \\
            \hline
            30 &      dh &    F &     7.77 &    62.09 &    56.24 &     93.76 &       86.76 \\
            \hline
            31 &       z &    F &    10.53 &    69.29 &    50.35 &     86.96 &       81.62 \\
            \hline
            32 &       s &    F &    20.85 &    83.52 &    96.10 &     96.29 &       99.20 \\
            \hline
            \hline
            33 &      oy &    D &     1.91 &    88.97 &    72.24 &     95.82 &       95.82 \\
            \hline
            34 &      aw &    D &     2.03 &    70.83 &    25.00 &      2.78 &       32.87 \\
            \hline
            35 &      ow &    D &     5.95 &    66.15 &    51.74 &      8.37 &       50.84 \\
            \hline
            36 &      ey &    D &     6.36 &    80.52 &    76.92 &     11.54 &       75.43 \\
            \hline
            37 &      ay &    D &     6.65 &    82.39 &    81.10 &     53.40 &       85.68 \\
            \hline
            \hline
            38 &      dx &    C &     7.55 &    86.49 &    78.72 &     95.64 &       97.13 \\
            \hline
            39 &     sil &    C &   100.00 &    84.89 &    94.83 &     99.02 &       99.38 \\
            \hline
            \hline
            \multicolumn{8}{|c|}{Performance for clean data training and testing (full set of 39 classes): GMM: 72.5\%, NVP: 76.9\%, Glow: 76.3\%, Voting: 86.6\% } \\
            \hline
            \multicolumn{8}{|c|}{V - Vowels, SV- Semivowels, N - Nasals, D - Dipthongs, C - Closures, P - Plosives } \\
            \hline
    \end{tabular}}
    \label{tab:voting_results}
\end{table*}

\begin{table*}[!htbp]
    \centering
    \caption{Test accuracy (in \%) for clean and various noisy conditions. We compare the performance of the best performing model between GMM-HMM, NVP-HMM and Glow-HMM with respect to the performance of the voting based decision fusion. We use the notation GMM/NVP/Glow to represent the best performing model among the three models.}
    \setlength{\tabcolsep}{5pt}
    \resizebox{\textwidth}{!}{\begin{tabular}{|*{9}{c|} }
            \hline
            \multicolumn{9}{|c|}{Performance for clean data as a reference: GMM: 72.5, NVP: 76.9,  Glow: 76.3, Voting: 86.6} \\
            \hline \hline
            \multirow{3}{*}{Type of Noise}
            & \multicolumn{8}{c|}{SNR levels for different kinds of noises} \\
            \cline{2-9}
            & \multicolumn{2}{c|}{25dB} & \multicolumn{2}{c|}{20dB} & \multicolumn{2}{c|}{15dB} & \multicolumn{2}{c|}{10dB} \\
            \cline{2-9}
            & GMM / NVP / Glow & Voting & GMM / NVP / Glow & Voting & GMM / NVP / Glow & Voting & GMM / NVP / Glow & Voting\\
            \cline{1-9}
            \texttt{white} & 67.9  & 76.5  & 61.4  & 66.5  & 53.6  & 54.1  & 45.8  & 42.6   \\
            \hline
            \texttt{pink} & 69.4  & 80.1  & 62.9 & 71.9  & 54.9  & 59.9  & 45.9  & 45.4 \\
            \hline
            \texttt{babble} & 72.4  & 83.2  & 68.9  & 79.7  & 63.1  & 71.7  & 53.8 & 57.4  \\
            \hline
            \texttt{hfchannel} & 71.8 & 80.9  & 67.9  & 75.3  & 61.1  & 66.1 & 51.9  & 53.8 \\
            \hline
            \hline
        \end{tabular}}
    \label{tab:cl_tr_no_te_table_all_voting}
\end{table*}


\section{Conclusions}
\label{sec:conclusion}

In pursuit of explainability under maximum-likelihood classification and optimization, we show that it is possible to combine modern neural networks (normalizing flows), modern computational tools (Kaldi, Python and PyTorch), time-tested generative models for sequential data (HMMs), their machine learning based optimization (EM and mini-batch gradient search), and knowledge in domain specific signal processing (MFCCs in speech signal processing). The output of our endeavor can provide state-of-the-art performance.

We deliberately refrain from using a discriminative approach in this article on the ground of explainability. For the same reason, we also avoid carefully crafted discriminatve features. Still we are able to achieve $86.6\%$ phone recognition accuracy whereas the latest result is $86.2\%$ \cite{Ravanelli2019}. We refrain to claim that we are better by $0.4\%$. Our experience is that reporting of experimental results in machine learning with real data has some undue opaqueness and published results are typically in favour of authors, even though a reproducible research culture has been steadily growing.





\section*{Acknowledgment}
The authors would like to thank KTH Digital Futures Center for support amid a pandemic.

\ifCLASSOPTIONcaptionsoff
  \newpage
\fi



%


\bibliographystyle{IEEEtran}
\bibliography{strings,refs}
%








\end{document}